\documentclass{article}

\usepackage{arxiv}
\usepackage[square,compress,numbers]{natbib}

\usepackage[utf8]{inputenc} 
\usepackage[T1]{fontenc}    
\usepackage{hyperref}       
\hypersetup{
    colorlinks=false,
    linkcolor=black,
    filecolor=black,      
    urlcolor=black,
}
\usepackage{url}            
\usepackage{booktabs}       
\usepackage{amsfonts}       
\usepackage{nicefrac}       
\usepackage{microtype}      
\usepackage{lipsum}
\usepackage{graphicx}
\usepackage{amsmath}

\usepackage{multirow}
\usepackage{todonotes}

\usepackage{xspace}
\newcommand{\hlsfml}{\texttt{hls4ml}\xspace}

\newcommand{\AutoQKeras}{\textsc{AutoQKeras}\xspace}

\usepackage{doi}
\usepackage{footnote}
\newlength\cmsTabSkip

\title{
Real-time semantic segmentation on FPGAs for autonomous vehicles with hls4ml}

\author{
   Nicol\`{o} Ghielmetti\thanks{Also at Politecnico di Milano, Italy}, Vladimir Loncar\thanks{Also at Institute of Physics Belgrade, Serbia}, Maurizio Pierini, Marcel Roed\thanks{Also at University of Oxford, UK}, Sioni Summers \\
  European Organization for Nuclear Research (CERN) \\
  CH-1211 Geneva 23, Switzerland
  \And
   Thea Aarrestad \\
  Institute for Particle Physics and Astrophysics, ETH Zürich\\
  8093 Zürich, Switzerland
  \And
  Christoffer Petersson\thanks{Also at Chalmers University of Technology, Sweden}\\
  Zenseact \\
  Gothenburg, 41756, Sweden
  \And
  Hampus Linander\\
  University of Gothenburg \\
  Gothenburg, 40530, Sweden
    \And
    Jennifer Ngadiuba\\
    Fermi National Accelerator Laboratory\\
    Batavia, IL 60510, USA
  \And 
  Kelvin Lin\thanks{Currently at Amazon, USA}\\
  University of Washington\\
  Seattle, WA 98195, USA 
  \And 
  Philip Harris\\
  Massachusetts Institute of Technology\\
  Cambridge, MA 02139, USA  
 
}

\newcommand{\tmop}[1]{\ensuremath{\operatorname{#1}}}

\begin{document}
\maketitle

\begin{abstract}
In this paper, we investigate how field programmable gate arrays can serve as hardware accelerators for real-time semantic segmentation tasks relevant for autonomous driving. Considering  compressed versions of the ENet convolutional neural network architecture, we demonstrate a fully-on-chip deployment with a latency of 4.9 ms per image, using less than 30\% of the available resources on a Xilinx ZCU102 evaluation board. The latency is reduced to 3 ms per image when increasing the batch size to ten, corresponding to the use case where  
the autonomous vehicle receives inputs from multiple cameras simultaneously. We show, through aggressive filter reduction and heterogeneous quantization-aware training, and an optimized implementation of convolutional layers, that the power consumption and resource utilization can be significantly reduced while maintaining  accuracy on the Cityscapes dataset. 
\end{abstract}


\section{Introduction}

Deep Learning has strongly reshaped computer vision in the last decade, bringing the accuracy of image recognition applications to unprecedented levels. Improved pattern recognition capabilities have had a significant impact on the advancement of research in science and technology. Many of the challenges faced by future scientific experiments, such as the CERN High Luminosity LHC~\cite{Apollinari:2284929} or the Square Kilometer Array observatory~\cite{ska}, and technological challenges faced by, for example, the automotive industry, will require the capability of processing large amounts of data in real-time, often through edge computing devices with strict latency and power-consumption constraints. This requirement has generated interest in the development of energy-effective neural networks, resulting in efforts like tinyML~\cite{banbury2021benchmarking}, which aims to reduce power consumption as much as possible without negatively affecting the model accuracy. 

Advances in Deep Learning for computer vision have had a crucial impact on the development of autonomous vehicles, enabling the vehicles to perceive their environment at ever-increasing levels of accuracy and detail. Deep Neural Networks are used for finding patterns and extracting relevant information from camera images, such as the precise location of the surrounding vehicles and pedestrians. In order for an autonomous vehicle to drive safely and efficiently, it must be able to react fast and make quick decisions. This imposes strict latency requirements on the neural networks that are deployed to run inference on resource-limited embedded hardware in the vehicle.  

In addition to algorithmic development, computer vision for autonomous vehicles has benefited from technological advances in parallel computing architecture~\cite{raina2009large}. The possibility of performing network training and inference on graphics processing units (GPUs) has made large and complex networks computationally affordable and testable on real-life problems. Due to their high efficiency, GPUs have become a common hardware choice in the automotive industry for on-vehicle Deep Learning inference.

In this paper, we investigate the possibility of exploiting field-programmable gate arrays (FPGAs) as a low-power, inference-optimized, highly parallelisable alternative to GPUs. By applying aggressive  filter-reduction and quantization of the model bit precision at training time, and by introducing a highly optimized firmware implementation of convolutional layers, we achieve the compression required to fit semantic segmentation models on FPGAs. We do so by exploiting and improving the \hlsfml library, which provides an automatic conversion of a given Deep Neural Network into C\texttt{++} code, which is given as input to a high level synthesis (HLS) library. The HLS library then translates this into FPGA firmware, to be deployed on hardware. Originally developed for scientific applications in particle physics that require sub-microsecond latency~\cite{Duarte:2018ite,Summers:2020xiy,Loncar:2020hqp,Iiyama:2020wap,Heintz:2020soy,Francescato:2021ezq,Sun:2022bxx}, \hlsfml has been successfully applied outside the domain of scientific research~\cite{Coelho:2020zfu,Aarrestad:2021zos}, specifically in the context of tinyML applications~\cite{Fahim:2021cic}.

Applying model compression at training time is crucial in order to minimize resource-consumption and maximize the model accuracy. To do so, we rely on quantization-aware training through the QKeras~\cite{qkeras} library, which has been interfaced to \hlsfml in order to guarantee an end-to-end optimal training-to-inference workflow~\cite{Coelho:2020zfu}.  

As a baseline, we start from the ENet~\cite{paszke2016enet} architecture, designed specifically to perform pixel-wise semantic segmentation for tasks requiring low latency operations. We modify the architecture, removing resource-consuming asymmetric convolutions, and dilated or strided convolutions. In addition, we apply filter ablation and quantization at training time. Finally, we optimize the implementation of convolutional layers in \hlsfml in order to significantly reduce the resource consumption. With these steps, we obtain a good balance between resource utilization and accuracy, enabling us to deploy the whole network on a Xilinx ZCU102 evaluation board \cite{zcu102}. 

This paper is organized as follows: 
The baseline dataset and model are described in sections~\ref{sec:data}~and~\ref{sec:model}, respectively. The model compression and the specific optimization necessary to port the compressed model to the FPGA are described in section~\ref{sec:compression}~and~\ref{sec:fpgaporting}. Conclusions are given in section~\ref{sec:conclusion}.




\section{Dataset}
\label{sec:data}

\begin{figure}[t!]
    \centering
    \includegraphics[width=0.48\textwidth]{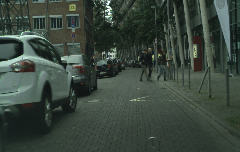}
    \includegraphics[width=0.48\textwidth]{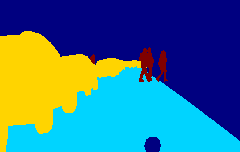}
    \caption{An downsampled image from the Cityscapes dataset (left) and the corresponding semantic segmentation target (right), in which the pixels belong to one of the classes  $\large\{$background (blue), road (teal), car (yellow), person (red)$\large\}$. }
    \label{fig:Cityscapes}
\end{figure}

Our experiments are performed using the Cityscapes dataset \cite{cordts2016cityscapes}, which involves 5,000 traffic scene images collected in 50 different cities with varying road types and seasons. These images have fine-grained semantic segmentation annotations with pixel-level classification labels. We have limited ourselves to the four semantic classes Road, Car, Person and Background. According to the standard Cityscapes split, 2975 images are used for training, 500 for validation and 1525  for testing. We crop and resize the original images to have an input resolution of 240$\times$152 pixels. As a pre-processing step, we normalize all pixel values (integer values in the range [0, 255]) to be in the [0, 1] range by dividing each one by 256. In this way all inputs are smaller than one and can be represented by a fixed-point datatype using only 8 bits ($\log_2 (256)$) (see section~\ref{sec:compression}). An example image from the dataset is shown in Fig.~\ref{fig:Cityscapes}, together with a visualization of its semantic segmentation mask. 

For evaluation metrics we use two typical figures of merit for semantic segmentation:
\begin{itemize}
    \item The model accuracy (Acc), defined as ${\rm Acc}=\frac{TP + TN}{TP + TN + FP +FN}$, where $TP$, $TN$, $FP$, and $FN$  are the fraction of true positives, true negatives, false positives, and false negatives, respectively.
    \item The mean of the class-wise Intersection over Union (mIoU), i.e., the average across classes of the Intersection-Over-Union (defined as ${\rm IOU} = \frac{TP}{TP + FP + FN}$).
\end{itemize}

\section{Baseline model}
\label{sec:model}

\begin{figure}[htb]
    \centering
    \includegraphics[width=0.4\textwidth]{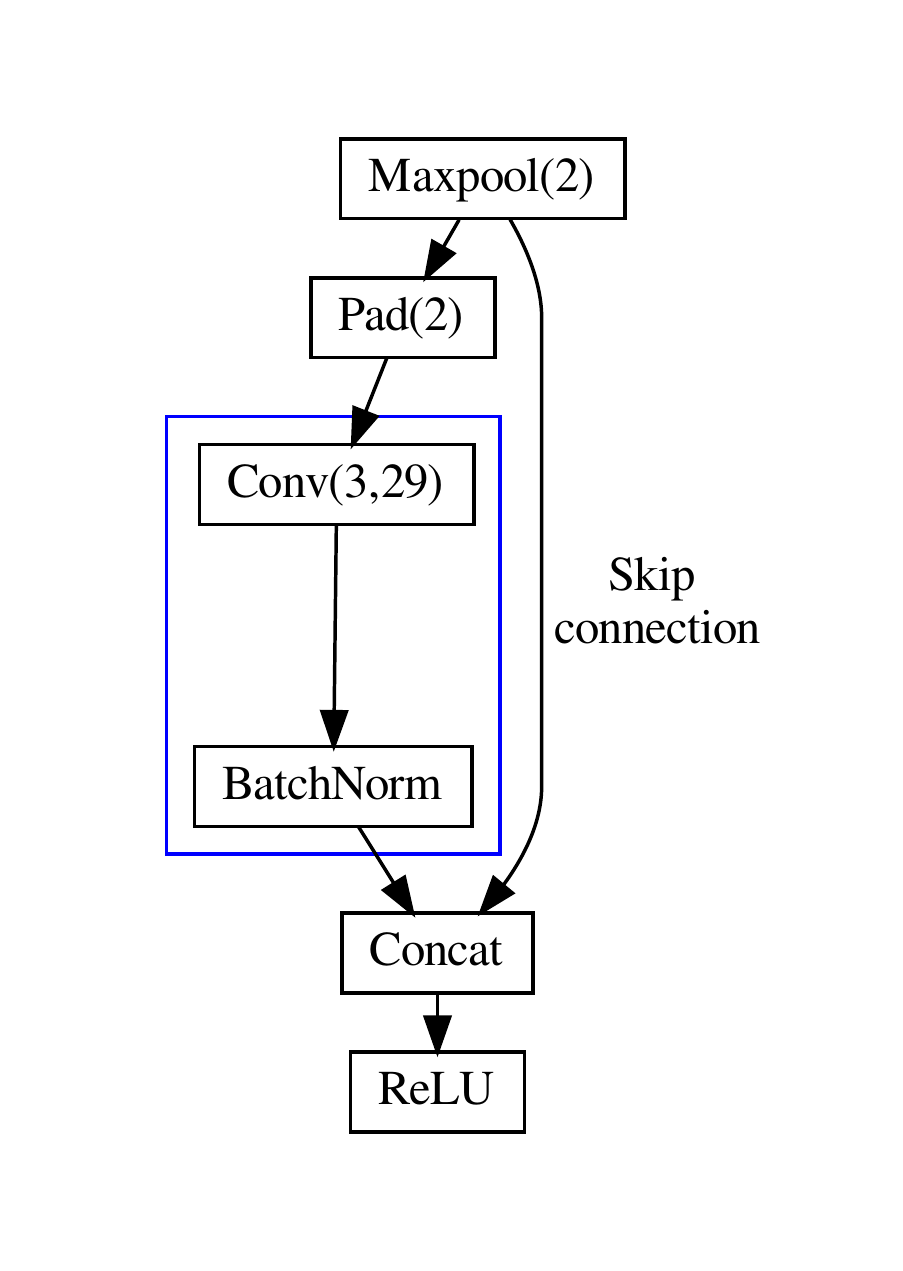} 
     \includegraphics[width=0.25\textwidth]{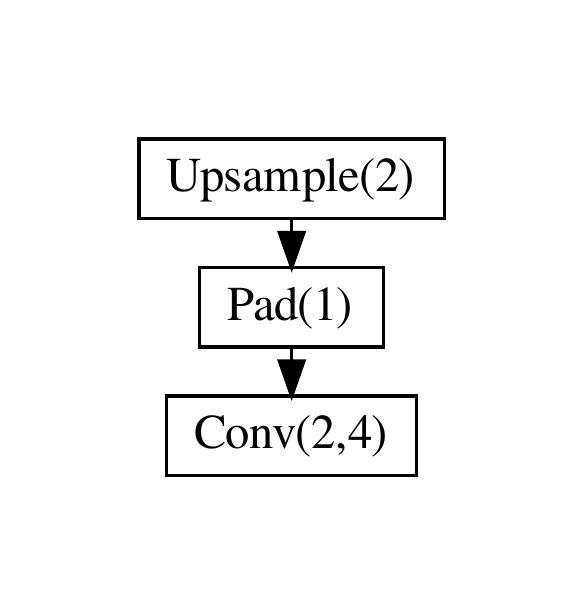}
   \caption{Initial (left) and final (right) block architecture. In the two diagrams, conv($k$, $f$) represents a convolutional layer with $f$ $k \times k$ filters; maxpool($k$) and upsample($k$) represent a $k \times k$ max pooling or upsample layer, respectively; and pad($p$) represents padding by $p$ pixels in the lower and right directions.\label{fig:initial}}
\end{figure}

\begin{figure}[htb]
    \centering
    \includegraphics[width=0.4\textwidth]{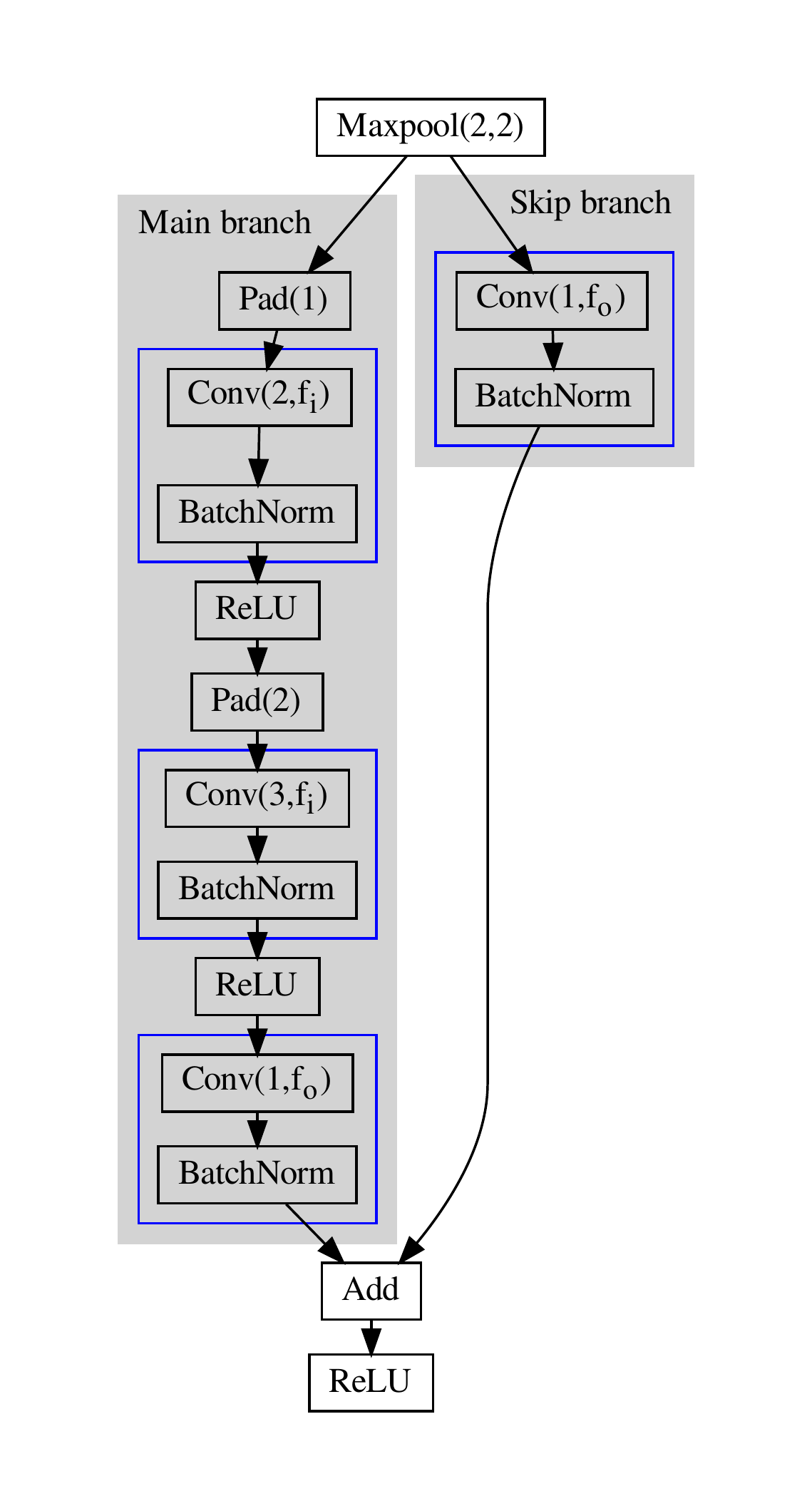} 
    \includegraphics[width=0.4\textwidth]{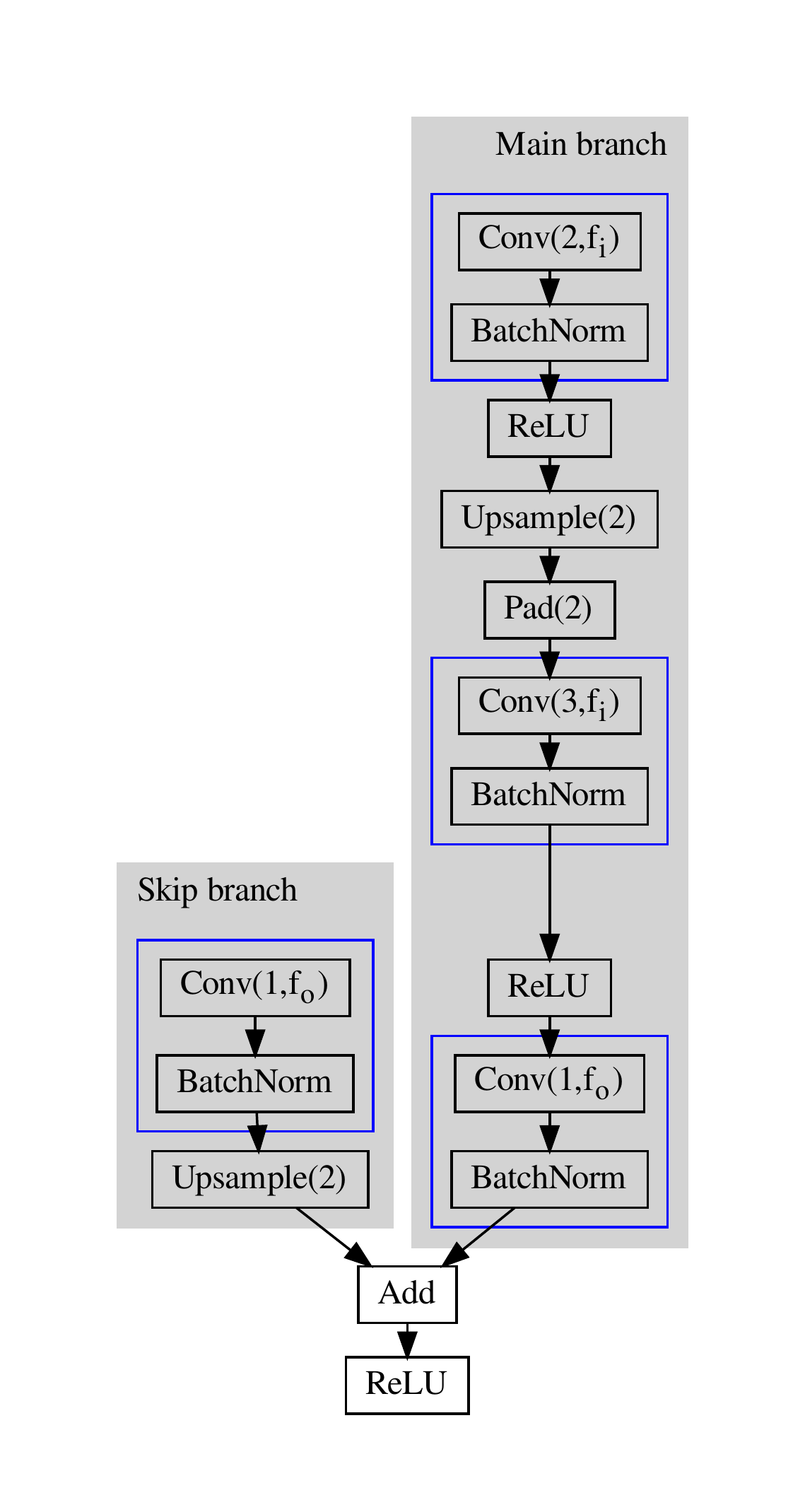} 
    \caption{Downsample encoder (left) and upsample decoder (right) blocks. In the figures, Conv($k$, $f$) represents a convolutional layer with $f$ $k \times k$ filters, Maxpool($k$) represents a $k \times k$ max pooling layer, Upsample($k$) represents a $k \times k$ upsampling layer, and Pad($p$) represents padding by $p$ pixels in the lower and right directions. Blue boxes represent convolutional and batch normalization layers that in the model used are single merged layers.\label{fig:blocks}}
\end{figure}

The architecture we use is inspired by a fully convolutional residual network called Efficient Neural Network (ENet)~\cite{paszke2016enet}. This network was designed for low latency and minimal resource usage. It is designed as a sequence of blocks, summarized in Table~\ref{tab:architecture}. The initial block, shown in the left figure in Fig.~\ref{fig:initial}, encodes the input into a $32 \times 120 \times 76$ tensor, which is then processed by a set of sequential blocks of bottlenecks. The first three blocks constitute the {\it downsampling encoder}, where each block consists of a series of layers as summarized in the left diagram in Fig.~\ref{fig:blocks}. The final two blocks provide an {\it upsampling decoder}, as illustrated in the right diagram in Fig.~\ref{fig:blocks}. The final block is shown in the right diagram of Fig.~\ref{fig:initial}.
\begin{table}[htb]
    \centering
    \begin{tabular}{c|c|c}
         Layer &  Type & Output resolution \\
         \hline
         Initial                 & downsample & $f_0 \times 120 \times 76$ \\
         $3 \times$ bottleneck 1 & downsample & $f_1 \times 60 \times 38$ \\
         $3 \times$ bottleneck 2 & downsample & $f_2 \times 30 \times 19$ \\
         $3 \times$ bottleneck 3 &            &  $f_3 \times 30 \times 19$ \\
         $3 \times$ bottleneck 4 & upsample   &   $f_4 \times 60 \times 38$ \\
         $3 \times$ bottleneck 5 & upsample   &  $f_5 \times 120 \times 76$ \\
         Final                   & upsample   &  $4 \times 240 \times 152$ \\
         \hline
    \end{tabular}
    \caption{Model architecture parametrized by the number of filters in the bottlenecks $f_i$, with $i=1,...,5$.\label{tab:architecture}}
\end{table}

Some differences from the original architecture in \cite{paszke2016enet} is that we do not use asymmetric, dilated, or strided convolutions. To further reduce the resource usage, we use three bottlenecks per block instead of five, and we merge convolutional layers with batch normalization layers by rescaling convolutional filter weights with batch normalization parameters (implemented through a \texttt{QConv2DBatchnorm} layer \cite{qc2dbn}). When we use quantization-aware training, this allows us to directly quantize the merged weights during the forward pass, rather than quantizing the batch normalization parameters and the convolutional filters separately. This merging of layers saves resources on the FPGA, since only the merged weights are used. Performing the merging already during training, ensures that the weights used during training and during inference are quantized the same way.
The baseline ENet model is obtained fixing the six $f$ hyperparameters of Table~\ref{tab:architecture} to $(32, 64, 64, 64, 128, 48)$. This choice results in an architecture with $1.1 \cdot 10^6$ parameters, yielding a mIoU=$63.2\%$ and an accuracy of $91.5\%$. 

\section{Model compression}
\label{sec:compression}

We consider two compression techniques for the model at hand: filter-wise homogeneous pruning, obtained by reducing the number of filters on all the convolutional layers; and quantization, i.e., reducing the number of bits allocated for the numerical representation of the network components and the output of each layer computation. 

In addition, we use the AutoQKeras library~\cite{Coelho:2020zfu}, distributed with QKeras, to optimize the numerical representation of each component at training time as a hyperparameter. This is done using a mathematical model of the inference power consumption as a constraint in the loss function.

\subsection{Filter multiplicity reduction}

  \begin{table}[hb]
  \centering
  \begin{tabular}{c|ccccccc|cc}
    Model name & $f_i$ & $f_1$ & $f_2$ & $f_3$ & $f_4$ & $f_5$ & $\tmop{Parameters}$ &
    $\tmop{mIoU}$(\%) & $\tmop{Accuracy}$(\%)\\
    \hline
    Enet & $32$ & $64$ & $64$ & $64$ & $128$ & $48$ & $1.1 \cdot 10^6$ & $63.2$ &
    $91.5$\\
    \hline
    {\bf Enet16} & $\mathbf{32}$ & $\mathbf{16}$ & $\mathbf{16}$ & $\mathbf{16}$ & $\mathbf{16}$ & $\mathbf{16}$ & $\mathbf{5 \cdot 10^4}$ & $\mathbf{54.3}$ &
    $\mathbf{87.9}$\\
    Enet12 & $32$ & $12$ & $12$ & $12$ & $12$ & $12$ & $3 \cdot 10^4$ & $52.0$ &
    $86.8$\\
    {\bf Enet8} & $\mathbf{32}$ & $\mathbf{8}$ & $\mathbf{8}$ & $\mathbf{8}$ & $\mathbf{8}$ & $\mathbf{8}$ & $\mathbf{1.4 \cdot 10^4}$ & $\mathbf{49.4}$ & $\mathbf{85.6}$\\
    Enet6 & $32$ & $6$ & $6$ & $6$ & $6$ & $6$ & $9 \cdot 10^3$ & $45.9$ & $84.0$\\
    {\bf Enet4} & $\mathbf{32}$ & $\mathbf{4}$ & $\mathbf{4}$ & $\mathbf{4}$ & $\mathbf{4}$ & $\mathbf{4}$ & $\mathbf{5 \cdot 10^3}$ & $\mathbf{36.6}$ & $\mathbf{81.5}$\\
    \hline
  \end{tabular}
  \caption{Architecture reduction through internal filter ablation and corresponding performance. As a reference, the baseline architecture is reported on the first row. Highlighted in bold the three models considered further in this work.\label{tab:enet_small_fp}}
\end{table}

Normally, network pruning consists of zeroing specific network parameters that have little impact on the model performance. This could be done at training time or after training. In the case of convolutional layers, a generic pruning of the filter kernels would result in sparse kernels. It would then be difficult to take advantage of pruning during inference. To deal with this, filter ablation (i.e., the removal of an entire kernel) was introduced~\cite{girshick2014rich}. When filter ablation is applied, one usually applies a restructuring algorithm (e.g., Keras Surgeon~\cite{kerassurgeon}) to rebuild the model into the smaller-architecture model that one would use at inference. In this work, we take a simpler (and more drastic) approach: we treat the number of filters in the convolutional layers as a single hyperparameter, fixed across the entire network. We then reduce its value and repeat the training, looking for a good compromise between accuracy and resource requirements.

We repeat the procedure with different target filter multiplicities. The result of this procedure is summarized in Table~\ref{tab:enet_small_fp}, where different pruning configurations are compared to the baseline Enet model. 

Out of these models, we select two configurations that would be affordable on the FPGA at hand: a four-filters (Enet4) and an eight-filter (Enet8) configuration. As a reference for comparison, we also consider one version with 16 filters, Enet16, despite it being too large to be deployed on the FPGA in question. We then proceed by quantizing these models through quantization-aware training to further reduce the resource consumption.

\subsection{Homogeneous quantization-aware training}

Homogeneous quantization-aware training (QAT) consists of repeating the model training while forcing the numerical representation of its weight and activation functions to a fixed $\langle T, I \rangle$ precision, where $T$ is the total number of bits and $I$ is the number of integer bits. Doing so, the model training converges to a minimum that might not be the absolute minimum of the full-precision training, but that would minimize the performance loss once quantization is applied.

In practice, we perform a homogeneous QAT replacing each layer of the model with its QKeras equivalent and exploiting the existing QKeras-to-\hlsfml interface for FPGA deployment.

We study the impact of QAT for $T\in{2, 4, 8}$ with $I=0$, on the pruned models desribed above (Enet4, Enet8, and Enet16). The resulting performance is shown in Table~\ref{tab:hom_quant}, where we label the three quantization configurations as Q2, Q4, and Q8, respectively.

The resulting resource utilization for Enet4 and Enet8 falls within the regime of algorithms that we could deploy on the target FPGA. We observe similar drops in accuracy when going from full precision to Q8 and from Q4 to Q2, but little differences between the Q4 and Q8 models. In this respect, Q4 would offer a better compromise between accuracy and resources than Q8.

\begin{table}[h!]
\centering
\begin{tabular}{c|c|cccccc|cc}
Model name & Quantization & $f_i$ & $f_1$ & $f_2$ & $f_3$ & $f_4$ & $f_5$  & mIoU (\%)  & Accuracy (\%) \\
\hline
Enet16 & - & 32 & 16 & 16 & 16 & 16 & 16 & 54.3 & 87.9\\
Enet8  & - & 32 &  8 &  8 &  8 &  8 & 8 & 49.4 & 85.6 \\
Enet4  & - & 32 &  4 &  4 &  4 &  4 & 4 & 36.6 & 81.5\\
\hline
Enet16Q8 & 8 & 32 & 16 & 16 & 16 & 16 & 16 & 35.0 & 79.1\\
\bf{Enet8Q8} & \bf{8} & \bf{32} &  \bf{8} &  \bf{8} &  \bf{8} &  \bf{8} &  \bf{8} &  \bf{33.4} & \bf{77.1}\\
Enet4Q8 & 8 & 32 & 4 & 4 & 4 & 4 & 4 & 13.6 & 53.8\\
Enet16Q4 & 4 & 32 & 16 & 16 & 16 & 16 & 16 & 34.1 & 77.9\\
\bf{Enet8Q4} & \bf{4} & \bf{32} & \bf{8} & \bf{8} & \bf{8} & \bf{8} & \bf{8} & \bf{33.9} & \bf{77.6}\\
Enet4Q4 & 4 & 32 & 4 & 4 & 4 & 4 & 4 & 13.5 & 53.6\\
Enet16Q2 & 2 & 32 & 16 & 16 & 16 & 16 & 16 & 27.4 & 68.6\\
Enet8Q2 & 2 & 32 & 8 & 8 & 8 & 8 & 8 & 28.7 & 71.1\\
Enet4Q2 & 2 & 32 & 4 & 4 & 4 & 4 & 4 & 13.4 & 53.5 \\
\hline
\bf{EnetHQ} & \bf{Heterogeneous} & \bf{8} & \bf{2} & \bf{4} & \bf{8} & \bf{4} & \bf{3} & \bf{36.8} & \bf{81.1} \\
\hline

\end{tabular}
\caption{Homogeneously and heterogeneously quantized models with indicated bitwidth and filter architecture together with their validation mean IOU trained with quantization aware training using QKeras. The corresponding values before quantization (from Table~\ref{tab:enet_small_fp}) are also reported in the three first rows.
\label{tab:hom_quant}}
\end{table}

Out of these, the models with the highest accuracy and mIoU that would be feasible to fit on the FPGA, is the 8 filter model quantized to 8 bits (Enet8Q8) and the 8 filter model quantized to 4 bits (Enet8Q4).

The quantization of the model does not have to be homogeneous across layers. In fact, it has been demonstrated that a heterogeneous quantization is the best way to maintain high accuracy at low resource-cost~\cite{qkeras_hls4ml}. We therefore define one final model with an optimized combination of quantizers. 

\subsection{Heterogeneous quantization aware training}

Heterogeneous QAT consists in applying different quantization to different network components. For deep networks, one typically deals with the large number of possible configurations by using an optimization library. In our case, we use \AutoQKeras~\cite{Coelho:2020zfu}. In \AutoQKeras, a hyperparameter search over individual layer quantization conditions and filter counts is performed. Since the model contains skip connections, the scan over number of filters needs to be handled with care. In particular, we use the block features of \AutoQKeras to ensure that the filter count matches throughout a bottleneck, so that the tensor addition of the skip connection will have valid dimensions.

The search for best hyperparameters, including the choice of indivdual quantizers for kernels and activations, is carried out using a Bayesian strategy where the balance between accuracy and resource usage is controlled by targeting a metric derived from them both\cite{Coelho:2020zfu}. In our search we permit e.g. a 4\% decrease in accuracy if the resource usage also is halved at the same time.


\begin{figure}
    \centering
    \includegraphics[width=0.95\textwidth]{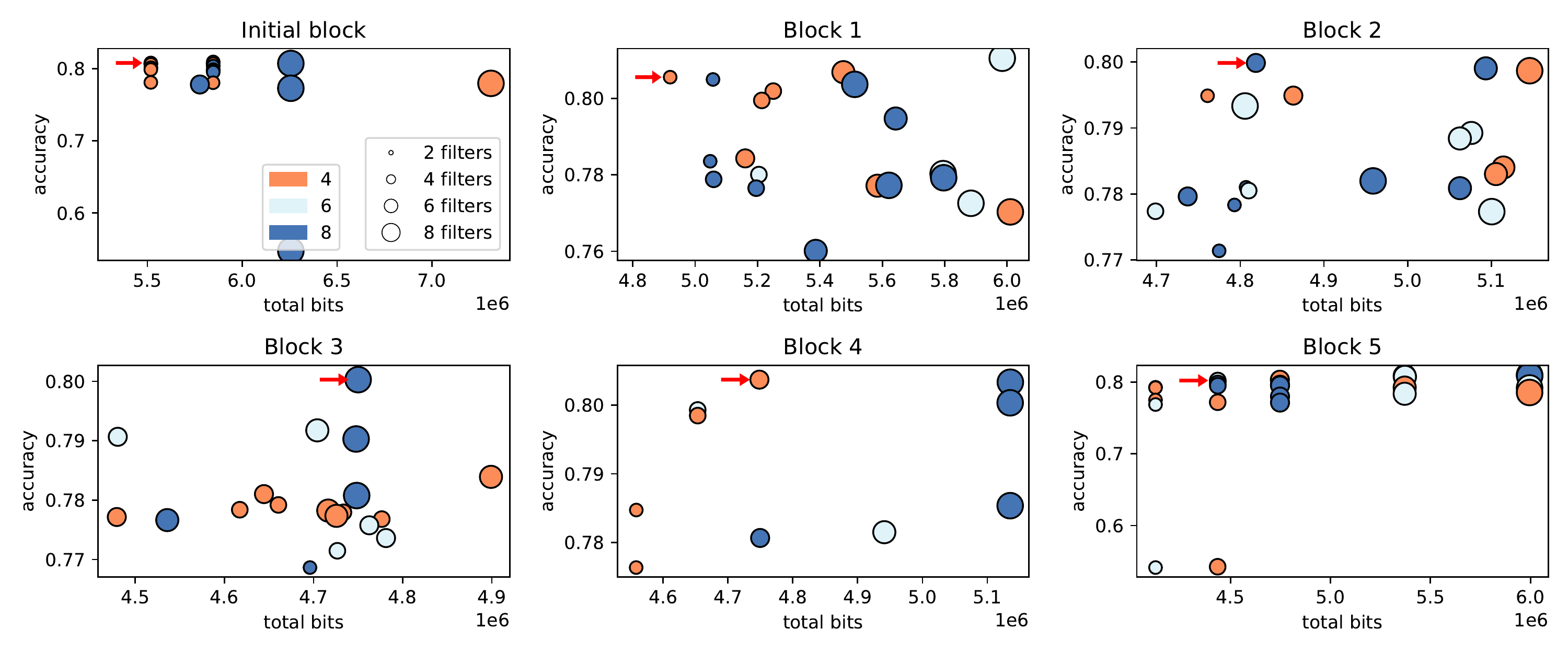}
    \caption{Validation accuracy versus number of bits for the models tested during heterogeneous QAT with \AutoQKeras, for the six blocks in the network. The circle size represents the number of filters, while the color refers to the median bits for the convolutions in the corresponding block. Details on the grid of options considered in the optimization are given in the text.\label{fig:autoq_i4_i8_f4}}
\end{figure}

The hyperparameter scan is done sequentially over the blocks, i.e. the Bayesian search over quantization and filter count of the initial layer is performed first and is then frozen for the hyperparameter scan of the first bottleneck and so on. The rest of the model is kept in floating point until everything in the end is quantized.

Figure~\ref{fig:autoq_i4_i8_f4} shows the outcome of the heterogeneous QAT, in terms of validation accuracy and total number of bits for the six blocks in the network. The optimal configuration search is performed taking as a baseline the Enet4 model, scanning the kernel bits in $\{4, 8\}$ and fixing the number of kernels to four times a by-layer multiplicative chosen in $\{0.5, 0.75, 1.0, 1.25, 1.5, 1.75, 2.0\}$. The optimal configuration (EnetHQ) is obtained for $f_i=8$, $f_1=2$, $f_2=4$, $f_3=8$, $f_4=4$, and $f_5=3$, resulting in $4.7 \cdot 10^3$ parameters, a mIoU=36.8\% and an accuracy of 81.1\%. Out of all the quantized models, both homogeneous and heterogeneous, this is the one which performs the best.


\section{FPGA implementation, deployment and results}
\label{sec:fpgaporting}

\subsection{Resource-efficient convolution algorithm}
\label{sec:line_buffer_CNN}

The \hlsfml library has an implementation of convolutional layers that is aimed at low-latency designs \cite{Aarrestad:2021zos}. However, this implementation comes at the expense of high resource utilization. This is due to the number of times pixels of the input image are replicated to maintain the state of a sliding input window. For convolutional layers operating on wider images, like in our case, this overhead can be prohibitively large. In order to reduce the resource consumption of the convolutional layers of the model, we introduce a new algorithm that is more resource efficient.

The new implementation, dubbed "line buffer", uses shift registers to keep track of previously seen pixels. The primary advantage of the line buffer implementation over the previous one is the reduction of the size of the buffer needed to store the replicated pixels. For an image of size $H \times W$, with a convolution kernel of size $K \times L$, the line buffer allocates $K-1$ buffers (chain of shift registers) of depth $W$ for the rows of the image, while the previous implementation allocates $K^2$ buffers of depth $K \times (W-K+1)$ for the elements in the sliding input window.

\begin{figure}[t!]
    \centering
    \includegraphics[width=0.70\textwidth]{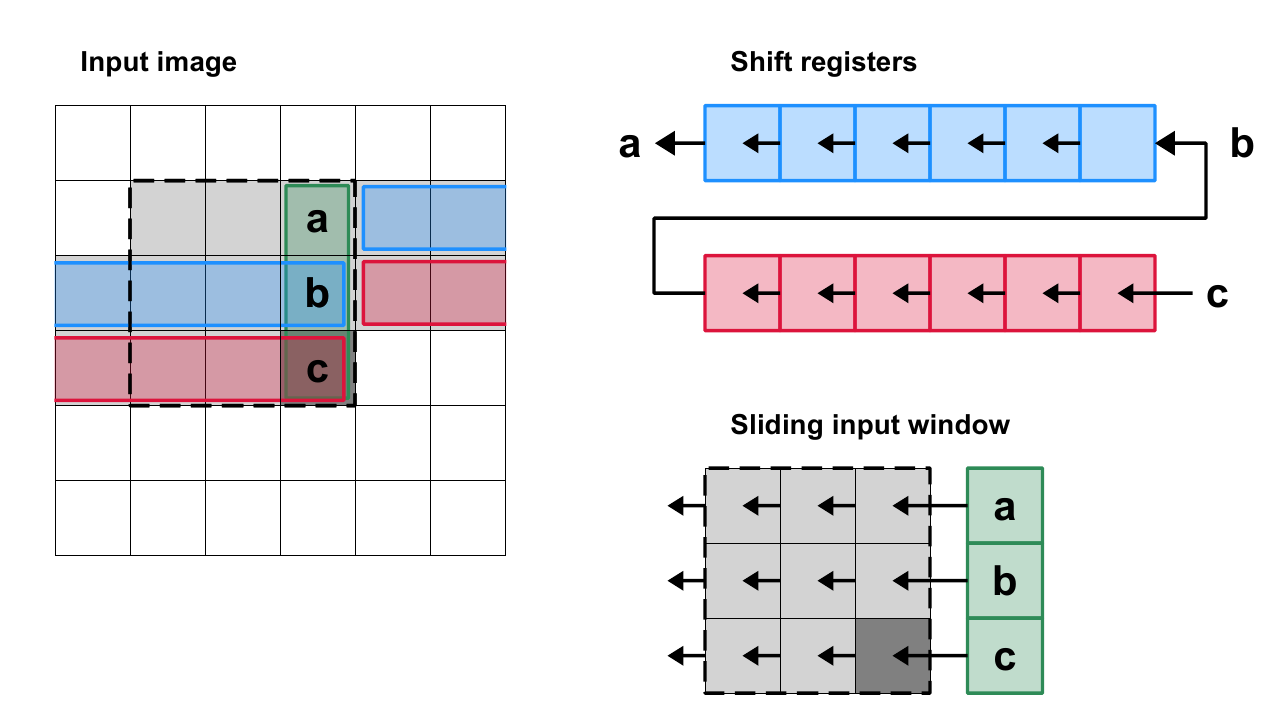}
    \caption{Schematic representation of the new \hlsfml implementation of Convolutional layers, as described in the text.\label{fig:line_buf_algo}}
\end{figure}

The algorithm is illustrated on Fig.~\ref{fig:line_buf_algo}. Initially, each new pixel read from the input image stream is pushed into the shift register
chain. If the shift register is full, the first element will be popped and it will be pushed into the next shift register in chain. The process is repeated for all $K-1$ shift registers in the chain. The popped pixels are stacked with the input pixel into a column vector and are pushed as the rightmost column of the input window. The pixels popped from the leftmost column of the input window are not used further. In our implementation, the propagation of new pixels through the shift register chain and the insertion into the sliding input window are completed in a single clock cycle, making the implementation as efficient as the existing \hlsfml implementation.

To compute the output from the populated sliding input window, we rely on the existing routines of \hlsfml. We rely on a set of counters to keep track of input window state to know when to produce an output. The algorithm for maintaining the chain of shift registers and populating the sliding input window can be adapted for use in the pooling layers as well.

To compare the two implementations, we consider the resource utilization of an ENet bottleneck block consisting of 8 filters, implemented using either method. The results are summarized in Table~\ref{tab:line_buffer_comp}. We observe a substantial reduction in BRAM usage, at the price of a small increase in LUT utilization.

\begin{table}[h!]
\centering
\begin{tabular}{c|cccc}
Implementation & BRAM & DSP & FF & LUT \\
\hline
Encoded \cite{Aarrestad:2021zos}    & 4,752 & 5,632 & 195,344 & 291,919 \\
Line buffer & 4,064 & 5,632 & 176,620 & 305,494 \\
\hline
Improvement & -15\% & 0\% & -1\% & +5\% \\
\hline
\end{tabular}
\caption{Comparison of previous and proposed \hlsfml implementation of the convolutional layer, in terms of relative reduction of resource utilization. The estimates are made targeting an xczu9eg-2ffvb1156 MPSoC device on a ZCU102 development kit.}
\label{tab:line_buffer_comp}
\end{table}

\subsection{FIFO depth optimization}
\label{sec:bram_CNN}

With the dataflow compute architecture of \hlsfml, layer compute units are connected with FIFOs, implemented as memories in the FPGA. 
These FIFOs contribute to the overall resource utilisation of the design.
The read and write pattern of these FIFOs depends on the dataflow through the model, which is not predictable before the design has been scheduled by the HLS compiler, and is generally complex.
With previous \hlsfml releases, these memories have therefore been assigned a depth corresponding to the dimensions of the tensor in the model graph as a safety precaution.

To optimize this depth and thereby reduce resource consumption, we implemented an additional step in the compilation of the model to hardware. By using the clock-cycle accurate RTL simulation of the scheduled design, we can monitor the actual occupancy of each FIFO in the model when running the simulation over example images. This enables us to extract and set the correct size of the FIFOs, reducing memory usage compared to the baseline.

By applying this procedure, we observe a memory efficiency 
$\dfrac{\sum_{l}O_l}{\sum_{l} F_l}=19.5\%$, where the index $l$ runs across the layers, $O_l$ is the observed occupancy for the $l$-th layer, and $F_l$ is the corresponding FIFO depth. The corresponding mean occupancy is found to be $\sum_{l}\dfrac{O_l}{F_l}=4.7\%$.

We then resize every FIFO to its observed maximum occupancy and rerun the C-Synthesis, thereby saving FPGA resources and allowing larger models to fit on the FPGA.
Table \ref{tab:fifo_depth_opt} shows the impact of such an optimization on the FPGA resources for one example model, Enet8Q8, demonstrating a significant reduction of resources.

  \begin{table}[h!]
    \centering
    \begin{tabular}{c|ccccc}
    Optimisation & BRAM & LUT & FF & DSP & Latency \\
    \hline
    No & 7,270 & 676,760 & 230,913 & 228 & 3.577 ms \\
    Yes & 1,398 & 437,559 & 146,392 & 228 & 3.577 ms \\ 
    \hline
    Improvement & -81\%  & -35\% & -37\%  & 0\%  & 0\%  \\
    \hline
    \end{tabular}
    \caption{Effect of FIFO depth optimization on FPGA resource usage and model latency. The values in the table are taken from Vivado HLS {\it estimates} of resource usage. A comparison using physical resource usage is unfeasible since the model without optimization can not be synthesized. The estimates are made targeting an xczu9eg-2ffvb1156 MPSoC device on a ZCU102 development kit.}
    \label{tab:fifo_depth_opt}
  \end{table}

\subsection{Results}
\label{sec:dse_results}
The hardware we target is a Zynq UltraScale+ MPSoC device (xczu9eg-2ffvb1156) on a ZCU102 development kit, which targets automotive applications.
After reducing the FPGA resource consumption through the methods described above, the highest accuracy models highlighted in Table~\ref{tab:hom_quant} are synthesized. These are the homogeneously quantized Enet8Q8 and Enet8Q4 models, as well as the heterogeneously quantized EnetHQ model. To find the lowest latency implementation, we run several attempts varying the {\it reuse factor} (RF) and the clock period. The RF indicates how many times a multiplier can be reused (zero for a fully parallel implementation). Lower RF leads to lower latency, but higher resource usage. We targeted reuse factors of 10, 20, 50, 100, and clock periods of 5, 7, 10 ns. For each model, we then chose the configuration yielding the lowest latency. For Enet8Q8, this is a target clock period of 7 ns and RF=10. For Enet8Q4 and EnetHQ we use a clock period of 7 ns and RF = 6.

Inference performance of this model was measured on the ZCU102 target device.
The final latency and resource utilization report is shown in Table~\ref{tab:fifo_depth_util}.
\begin{table}[h!]
    \centering \small
    \label{tab:fifo_depth_util}
     \caption{Accuracy, mIoU, latency and resource utilization for the EnetHQ, Enet8Q4 and Enet8Q8 models. The latency is quoted for a batch size $b=1$ and $b=10$. Resources are expressed as a percentage of those available on the xczu9eg-2ffvb1156 MPSoC device on the ZCU102 development kit. The last row is a comparison to work presented in Ref.~\cite{enetOnFpga}.}
    \begin{tabular}{c|c|c|cc|c|c|c|c}
        Model & Acc. & mIoU & \multicolumn{2}{|c|}{Latency [ms]} & BRAM & LUT & FF & DSP \\
                 & &     & b=1 & b=10 &  &  &  \\
        \hline
        EnetHQ  & 81.1\% & 36.8 \% & 4.9 & 30.6 & 224.5 (25\%) &  76,718 (30\%) & 87,059 (16\%) & 450 (18\%)\\
        Enet8Q4 & 77.6\% & 33.9 \% & 4.8 & 30.2  &  342.0 (37\%)& 166,741 (61\%)& 90,536 (16\%)& 0\\
        Enet8Q8 & 77.1\% & 33.4 \% & 4.8 & 30.0 & 508.5 (56\%) & 126,458 (46\%) & 134,385 (25\%) & 1,502 (60\%)\\
        \hline
        ENet~\cite{enetOnFpga}   & -     & 63.1\%  &30.38 (720)$^a$& -    & 257          & 62,599         & 192,212        & 689 \\
    \end{tabular}
    \vspace{2mm}
    \raggedright
    \footnotesize{\newline$^a$ The former is without considering data transfer, pre- and post-processing. The number in parenthesis includes these additional overheads, averaged over 58 images, and is more comparable to the numbers we present.}\\
\end{table}

We measured the time taken by the accelerator to produce a prediction on batches of images, with batch sizes of $b=1$ and $b=10$. The same predictions have been executed $10^5$ times, and the  time average is taken as the latency. The single image latency (batch size of 1) is 4.8-4.9 ms for all three models. Exploiting the data flow architecture, the latency to process images in a batch size of 10 is less than 10 times the the latency observed for a batch size of 1. While in a real-world deployment of this model the latency to return the predictions of a single image is the most important metric, a system comprised of multiple cameras may be able to benefit from the speedup of batched processing by batching over the images captured simultaneously from different cameras. The model with the highest accuracy and lowest resource consumption is the heterogeneously quantized EnetHQ model. This model has an mIoU of 36.8\% and uses less than 30\% of the total resources.

Similar work on performing semantic segmentation on FPGAs include Ref.~\cite{enetOnFpga} and a comparison is given in Table~\ref{tab:fifo_depth_util}. Here, the original ENet model~\cite{paszke2016enet} is trained and evaluated on the Cityscapes dataset, and then deployed on a Xilinx Zynq 7035 FPGA using the Xilinx Vitis AI Deep Learning Processor Unit (DPU). There are some crucial differences between the approach taken here and that of Ref.~\cite{enetOnFpga}. In order to achieve the lowest possible latency, we implement a fully on-chip design with high layer parallelism. We optimize for latency, rather than frame rate, such that in a real-life application the vehicle response time could be minimized. Keeping up with the camera frame rate is a minimal requirement, but a latency lower than the frame interval can be utilized. In our approach, each layer is implemented as a separate module and data is streamed through the architecture layer by layer. Dedicated per-layer buffers ensure that just enough data is buffered in order to feed the next layer. This is highly efficient, but limits the number of layers that can be implemented on the FPGA. Consequently, in order to fit onto the FPGA in question, our model is smaller and achieves a lower mIoU. Ref.~\cite{enetOnFpga} does not quote a latency, but a frame rate. A best-case latency is then computed as the inverse of this frame rate, which corresponds to 30.38 ms. However, this does not include any overhead latency like data transfer, pre- and post-processing. Including these, the average time per image increases to 720 ms.

\section{Conclusions}
\label{sec:conclusion}
In this paper, we demonstrate that we can perform semantic segmentation on a single FPGA on a Zynq MPSoC device using a compressed version of ENet. The network is compressed using automatic heterogeneous quantization at training time and a filter ablation procedure, and is then evaluated on the Cityscapes dataset. Inference is executed on hardware with a latency of $4.9$~ms per image, utilizing 18\% of the DSPs, 30\% of the LUTs, 16\% of the FFs and 25 \% of the BRAMs. 
Processing the images in batches of ten results in a latency of 30 ms per batch, which is significantly faster than ten times the single-image batch inference latency. This is relevant when batching over images captured from different cameras simultaneously. By introducing an improved implementation of convolutional layers in \hlsfml, we significantly reduce resource consumption, allowing for a fully-on-chip deployment of larger convolutional neural networks. This avoids latency overhead caused by data transfers between off-chip memory and FPGA processing elements, or between multiple devices. Also taking into account the favorable power-efficiency of FPGAs, we conclude that FPGAs offer highly interesting, low-power alternatives to GPUs for on-vehicle deep learning inference and other computer vision tasks requiring low-power and low-latency. 



\section*{Acknowledgment}

We acknowledge the Fast Machine Learning collective as an open community of multi-domain experts and collaborators. This community was important for the development of this project. 

M.~P., M.~R., S.~S., and V.~L. are supported by the European Research Council (ERC) under the European Union's Horizon 2020 research and innovation program (Grant Agreement No. 772369). M.~P. and N.~G. are supported by the European Research Council (ERC) under the European Union's Horizon 2020 research and innovation program (Grant Agreement No. 966696).

\appendix
\clearpage
\bibliographystyle{lucas_unsrt}   
\bibliography{references}

\providecommand{\href}[2]{#2}\begingroup\raggedright\begin{thebibliography}{10}%
\makeatletter
\providecommand{\hrefCMSnoop }[0]{\@secondoftwo}%
\makeatother
\providecommand{\doi}{\texttt{doi:}\begingroup \urlstyle{tt}\Url}

\bibitem{Apollinari:2284929}
A.~G. {et~al.}, ``{High-Luminosity Large Hadron Collider (HL-LHC): Technical
  Design Report V. 0.1}''.
\newblock CERN Yellow Reports: Monographs. CERN, Geneva, 2017.
\newblock
  \href{http://dx.doi.org/10.23731/CYRM-2017-004}{\doi{10.23731/CYRM-2017-004}}.

\bibitem{ska}
M.~A. Garrett\hrefCMSnoop {}{ {et~al.}, ``Square kilometre array: a concept
  design for phase 1'',} 2010.
\newblock
  \href{http://dx.doi.org/10.48550/ARXIV.1008.2871}{\doi{10.48550/ARXIV.1008.2871}},
  \url {https://arxiv.org/abs/1008.2871}.

\bibitem{banbury2021benchmarking}
C.~R. Banbury\hrefCMSnoop {}{ {et~al.}, ``Benchmarking tinyml systems:
  Challenges and direction'',} 2021.

\bibitem{raina2009large}
\hrefCMSnoop {}{R.~Raina, A.~Madhavan, and A.~Y. Ng, ``Large-scale deep
  unsupervised learning using graphics processors'',} in \textit{ Proceedings
  of the 26th annual international conference on machine learning},
  pp.~873--880, ACM.
\newblock 2009.

\bibitem{Duarte:2018ite}
\hrefCMSnoop {}{J.~Duarte {et~al.}, ``{Fast inference of deep neural networks
  in FPGAs for particle physics}'',} \textit{ J. Instrum.} \textbf{ 13} (2018),
  no.~07, P07027,
  \href{http://dx.doi.org/10.1088/1748-0221/13/07/P07027}{\doi{10.1088/1748-0221/13/07/P07027}},
  \href{http://www.arXiv.org/abs/1804.06913}{\texttt{arXiv:1804.06913}}.

\bibitem{Summers:2020xiy}
\hrefCMSnoop {}{S.~Summers {et~al.}, ``{Fast inference of Boosted Decision
  Trees in FPGAs for particle physics}'',} \textit{ J. Instrum.} \textbf{ 15}
  (2020), no.~05, P05026,
  \href{http://dx.doi.org/10.1088/1748-0221/15/05/P05026}{\doi{10.1088/1748-0221/15/05/P05026}},
  \href{http://www.arXiv.org/abs/2002.02534}{\texttt{arXiv:2002.02534}}.

\bibitem{Loncar:2020hqp}
\hrefCMSnoop {}{V.~Loncar {et~al.}, ``{Compressing deep neural networks on
  FPGAs to binary and ternary precision with HLS4ML}'',} \textit{ Mach. Learn.
  Sci. Tech.} \textbf{ 2} (2021) 015001,
  \href{http://dx.doi.org/10.1088/2632-2153/aba042}{\doi{10.1088/2632-2153/aba042}},
  \href{http://www.arXiv.org/abs/2003.06308}{\texttt{arXiv:2003.06308}}.

\bibitem{Iiyama:2020wap}
Y.~Iiyama\hrefCMSnoop {}{ {et~al.}, ``Distance-weighted graph neural networks
  on fpgas for real-time particle reconstruction in high energy physics'',}
  \textit{ Frontiers in Big Data} \textbf{ 3} (2021) 44,
  \href{http://dx.doi.org/10.3389/fdata.2020.598927}{\doi{10.3389/fdata.2020.598927}}.

\bibitem{Heintz:2020soy}
A.~Heintz\href
  {https://ml4physicalsciences.github.io/2020/files/NeurIPS_ML4PS_2020_137.pdf}{
  {et~al.}, ``Accelerated charged particle tracking with graph neural networks
  on {FPGAs}'',} in \textit{ {3rd Machine Learning and the Physical Sciences
  Workshop at the 34th Annual Conference on Neural Information Processing
  Systems}}.
\newblock 12, 2020.
\newblock
  \href{http://www.arXiv.org/abs/2012.01563}{\texttt{arXiv:2012.01563}}.

\bibitem{Francescato:2021ezq}
\hrefCMSnoop {}{S.~Francescato {et~al.}, ``{Model compression and
  simplification pipelines for fast deep neural network inference in FPGAs in
  HEP}'',} \textit{ Eur. Phys. J. C} \textbf{ 81} (2021), no.~11, 969,
  \href{http://dx.doi.org/10.1140/epjc/s10052-021-09770-w}{\doi{10.1140/epjc/s10052-021-09770-w}}.
  [Erratum: Eur.Phys.J.C 81, 1064 (2021)].

\bibitem{Sun:2022bxx}
C.~Sun\hrefCMSnoop {}{ {et~al.}, ``{Fast Muon Tracking with Machine Learning
  Implemented in FPGA}'',}
  \href{http://www.arXiv.org/abs/2202.04976}{\texttt{arXiv:2202.04976}}.

\bibitem{Coelho:2020zfu}
\hrefCMSnoop {}{C.~N. Coelho {et~al.}, ``{Automatic heterogeneous quantization
  of deep neural networks for low-latency inference on the edge for particle
  detectors}'',} \textit{ Nat Mach Intell} \textbf{ 3} (2021) 675–686,
  \href{http://dx.doi.org/10.1038/s42256-021-00356-5}{\doi{10.1038/s42256-021-00356-5}},
  \href{http://www.arXiv.org/abs/2006.10159}{\texttt{arXiv:2006.10159}}.

\bibitem{Aarrestad:2021zos}
\hrefCMSnoop {}{T.~Aarrestad {et~al.}, ``{Fast convolutional neural networks on
  FPGAs with hls4ml}'',} \textit{ Mach. Learn. Sci. Tech.} \textbf{ 2} (2021),
  no.~4, 045015,
  \href{http://dx.doi.org/10.1088/2632-2153/ac0ea1}{\doi{10.1088/2632-2153/ac0ea1}},
  \href{http://www.arXiv.org/abs/2101.05108}{\texttt{arXiv:2101.05108}}.

\bibitem{Fahim:2021cic}
\hrefCMSnoop {}{F.~Fahim {et~al.}, ``{hls4ml: An Open-Source Codesign Workflow
  to Empower Scientific Low-Power Machine Learning Devices}'',} in \textit{
  {tinyML Research Symposium 2021}}.
\newblock 3, 2021.
\newblock
  \href{http://www.arXiv.org/abs/2103.05579}{\texttt{arXiv:2103.05579}}.

\bibitem{qkeras}
\href {https://github.com/google/qkeras}{C.~Coelho {et~al.}, ``Qkeras'',} 2019.
\newblock \url {https://github.com/google/qkeras}.

\bibitem{paszke2016enet}
\hrefCMSnoop {}{A.~Paszke, A.~Chaurasia, S.~Kim, and E.~Culurciello, ``Enet: A
  deep neural network architecture for real-time semantic segmentation'',}
  2016.

\bibitem{zcu102}
\href
  {https://www.xilinx.com/products/boards-and-kits/ek-u1-zcu102-g.html#information}{``Xilinx
  ZCU102 evaluation board''.} \url
  {https://www.xilinx.com/products/boards-and-kits/ek-u1-zcu102-g.html#information}.

\bibitem{cordts2016cityscapes}
M.~Cordts\hrefCMSnoop {}{ {et~al.}, ``The cityscapes dataset for semantic urban
  scene understanding'',} 2016.

\bibitem{qc2dbn}
\href
  {https://github.com/google/qkeras/blob/master/qkeras/qconv2d_batchnorm.py}{``QConv2DBatchnorm
  layer implementation''.} \url
  {https://github.com/google/qkeras/blob/master/qkeras/qconv2d_batchnorm.py}.

\bibitem{girshick2014rich}
\hrefCMSnoop {}{R.~Girshick, J.~Donahue, T.~Darrell, and J.~Malik, ``Rich
  feature hierarchies for accurate object detection and semantic
  segmentation'',} 2014.

\bibitem{kerassurgeon}
\href {https://github.com/BenWhetton/keras-surgeon}{B.~Whetton, ``Keras
  surgeon'',} 2016.
\newblock \url {https://github.com/BenWhetton/keras-surgeon}.

\bibitem{qkeras_hls4ml}
C.~N. Coelho\hrefCMSnoop {}{ {et~al.}, ``Automatic heterogeneous quantization
  of deep neural networks for low-latency inference on the edge for particle
  detectors'',} \textit{ Nature Machine Intelligence} \textbf{ 3} (2021),
  no.~8, 675--686.

\bibitem{enetOnFpga}
\hrefCMSnoop {}{W.~Jia, J.~Cui, X.~Zheng, and Q.~Wu, ``Design and
  implementation of real-time semantic segmentation network based on fpga'',}
  in \textit{ 2021 7th International Conference on Computing and Artificial
  Intelligence}, ICCAI 2021, p.~321–325.
\newblock Association for Computing Machinery, New York, NY, USA, 2021.
\newblock
  \href{http://dx.doi.org/10.1145/3467707.3467756}{\doi{10.1145/3467707.3467756}}.

\end{thebibliography}\endgroup
\appendix

\end{document}